\documentclass[english,a4paper,11pt]{article}
\usepackage{fullpage}

\usepackage{pdflscape}
\usepackage[figuresright]{rotating} 
\usepackage{afterpage}
\usepackage{floatpag}
\usepackage{color}
\usepackage{array}
\usepackage{subcaption}
\usepackage{booktabs}

\usepackage{amsxtra, amsfonts, amssymb, amstext}
\usepackage{amsthm}
\usepackage{url}
\usepackage[noadjust]{cite}
\usepackage[colorlinks]{hyperref}
\definecolor{linkblue}{rgb}{0.1,0.1,0.8}
\hypersetup{colorlinks=true,linkcolor=linkblue,filecolor=linkblue,urlcolor=linkblue,citecolor=linkblue}

\newtheorem{theorem}{Theorem}
\newtheorem{definition}[theorem]{Definition}

\begin{document}

%
\title{Online Selection of CMA-ES Variants}

\author{Diederick Vermetten$^1$, Sander van Rijn$^1$, Thomas B{\"a}ck$^1$, Carola Doerr$^2$}
\date{$^1$Leiden Institute for Advanced Computer Science, Leiden, The Netherlands\\
$^2$Sorbonne University, LIP6, and CNRS, Paris, France}

\maketitle 

\begin{abstract}
In the field of evolutionary computation, one of the most challenging topics is algorithm selection. Knowing which heuristics to use for which optimization problem is key to obtaining high-quality solutions. We aim to extend this research topic by taking a first step towards a selection method for adaptive CMA-ES algorithms. We build upon the theoretical work done by van Rijn \textit{et al.} [PPSN'18], in which the potential of switching between different CMA-ES variants was quantified in the context of a modular CMA-ES framework.

We demonstrate in this work that their proposed approach is not very reliable, in that implementing the suggested adaptive configurations does not yield the predicted performance gains.
We propose a revised approach, which results in a more robust fit between predicted and actual performance. The adaptive CMA-ES approach obtains performance gains on 18 out of 24 tested functions of the BBOB benchmark, with stable advantages of up to 23\%. An analysis of module activation indicates which modules are most crucial for the different phases of optimizing each of the 24 benchmark problems. The module activation also suggests that additional gains are possible when including the (B)IPOP modules, which we have excluded for this present work.
\end{abstract}

\section{Introduction}
The creation of optimization algorithms has long been a topic of study in Mathematics and Computer Science. As the complexity of optimization problems increased, the need for fast heuristic algorithms became clear. Techniques like evolution strategies have been around for quite some time, and much research has been done into improving them, such as the popular Covariance Matrix Adaptation Evolution Strategy (CMA-ES)~\cite{hansen2001self_adaptation_es}. 

For CMA-ES, many adaptations have been developed over the last decade, all with the goal of creating a better optimization algorithm for certain types of problems. However, with all these different options to choose from, it can be difficult to know which version to use. The comparison of different CMA-ES variants consistently shows that there is no single best variant~\cite{BFK13,hansen2010comparing}, giving rise to the question how to best choose which of the variants to use for which kind of optimization problems.
In addition, all CMA-ES variants have several hyper-parameters, whose values can have considerable impact on their performance. This means that besides the \emph{algorithm selection problem} we also face an \emph{algorithm configuration problem}. These two problems are highly interlinked. The choice of algorithm and corresponding parameter configuration for a given problem at hand thereby forms itself a new optimization problem, which has gained increasing attention in the last decades. With the increased accuracy of performance prediction models,  state-of-the-art research in evolutionary computation focuses on \emph{automating} the decision of which algorithm and which parameter configuration to select.
Among the most widely used prediction models are Gaussian processes~\cite{bartz2010sequential,bartz2010experimental}, but also techniques from the machine learning community, such as random forests~\cite{hutter2011sequential,kerschke2017automated}. Surveys on algorithm selection are available in~\cite{munoz2015algorithm,kerschke2018survey,hoos2016automated_algorithm_selecion}.

While algorithm selection generally considers \emph{static} choices of algorithms, we aim to extend existing work to allow for \emph{adaptive} changes of the algorithm configuration. Note here that  \emph{on-the-fly} algorithm selection and configurations are also studied under the umbrella term   \emph{hyper-heuristics}~\cite{BurkeGHKOOQ13,burke2007graph}. To date, however, research on hyper-heuristics mostly focuses on discrete optimization and much less on continuous optimization. In particular, we are not aware of any previous approaches to address a dynamic selection of CMA-ES module variants.

We base our work on the modular CMA-ES framework introduced in~\cite{van_rijn_evolving_2016}. Within this framework, we can create many variants of CMA-ES by turning on or off certain modules, e.g. elitism, orthogonal sampling, and different weighting schemes (cf. Table~\ref{tab:es-opt}). Each module combination gives a different \emph{configuration} of the modular CMA-ES. 

Using this modular CMA-ES framework, a theoretical analysis of the potential benefits of adaptive CMA-ES configurations has been provided in~\cite{van_rijn_ppns_2018_adpative}. In this work, we will extend their study to all 24 BBOB functions~\cite{hansen_coco:_2016}, quantifying the actual performance gains, identify a need for a more reliable prediction, and propose an alternative method, which shows a much better fit between predicted performances and actual achieved running times. 

We focus on the case in which the CMA-ES variant can be changed only once. Our main results show that even such single switches allow for performance gains on 18 out of the 24 studied BBOB functions. In 11 cases the relative gain against the best static configurations are larger than 5\%, 8 functions show a relative gain  $\ge 10\%$ and the largest observed improvement (for F21) is 51\%. However, the 51\% is achieved for one particular configuration swap only, and when averaging the 10 best adaptive configuration pairs, the relative gain over the best static variant vanishes. We nevertheless see other functions with stable relative gains. The largest stable gains of around 20\%  are obtained for F5 and F6.

All numbers reported above are with respect to the default parameter-choices. That is, we focus in this work exclusively on the algorithms selection aspect, thereby leaving the question of configuring the default hyper-parameters for future work. Additional performance gains can be expected from such parameter tuning, as, for example, the results presented in~\cite{BelkhirDSS17} indicate.

Investigating the best-performing modules per each stage of the optimization process, we identify which configurations are most efficient for which parts of the optimization process. 

\textbf{Structure of the paper:}  Section~\ref{sec:modules} summarizes the modular framework, in which we build our adaptive CMA-ES. We introduce our implementation of the approach used in~\cite{van_rijn_ppns_2018_adpative} in Section~\ref{sec:implementing}, followed by a discussion of our additional techniques for configuration selection. Section~\ref{sec:experiments1} shows that the original approach is not very stable. In Section~\ref{sec:results} we discuss the results from our experiment, with a focus on the configurations and switching points at which the modules are changed. 
We conclude our contribution in Section~\ref{sec:conclusion} by looking at possible directions for future work.

\section{The Modular CMA-ES Framework}
\label{sec:modules}
The adaptive CMA-ES configurations we use were implemented in the modular CMA-ES framework  introduced in~\cite{van_rijn_evolving_2016}, which is freely available at~\cite{modCMA}. 
 
This framework implements 11 different modules. Of these 11 modules, 9 are binary and 2 are ternary, allowing for a combined total of $4,\hspace{-1pt}608$ different configurations. The full list of available modules is shown in Table~\ref{tab:es-opt}. 

To clarify the terminology, note that we follow the example given in~\cite{van_rijn_ppns_2018_adpative} and consider each combination of the 11 modules a ``configuration''. As mentioned in the introduction, we do not, in this work, change or optimize the hyper-parameters of the CMA-ES variants (such as the population size, the speed of the step size adaptation, etc.). We use the default values for these parameters instead.

\begin{table}
\small
    \renewcommand{\arraystretch}{1.1}
    \setlength\tabcolsep{5pt}

    \begin{tabular}{@{}lllll@{}}
    \toprule
    \# & Module name             & 0     & 1                 & 2 \\
    \midrule
    1  & Active Update~\cite{jastrebski_improving_2006}           & off              & on                & - \\
    2  & Elitism                 & ($\mu, \lambda$) & ($\mu{+}\lambda$) & - \\
    3  & Mirrored Sampling~\cite{brockhoff_mirrored_2010}     & off              & on                & - \\
    4  & Orthogonal Sampling~\cite{wang_mirrored_2014}     & off              & on                & - \\
    5  & Sequential Selection~\cite{brockhoff_mirrored_2010}    & off              & on                & - \\
    6  & Threshold Convergence~\cite{piad-morffis_evolution_2015}   & off              & on                & - \\
    7  & TPA~\cite{hansen_cma-es_2008}                     & off              & on                & - \\
    8  & Pairwise Selection~\cite{auger_mirrored_2011}      & off              & on                & - \\
    9  & Recombination Weights~\cite{auger2005quasi_random} & $\log(\mu{+}\frac{1}{2}){-}\frac{\log(i)}{\sum_j w_j}$ & $\frac{1}{\mu}$& - \\
    10 & Quasi-Gaussian Sampling & off             & Sobol            & Halton \\
    11 & Increasing Population~\cite{auger_restart_2005,hansen_benchmarking_2009}   & off             & IPOP             & BIPOP \\
    \bottomrule
    \end{tabular}
        \footnotesize
    \caption{Overview of the CMA-ES modules available in the used framework. 
    The entries in row 9 
    specify the formula for calculating each weight $w_i$.}    
    \label{tab:es-opt}
\end{table}

For our benchmarking, we use the BBOB-suite of benchmark functions~\cite{hansen_real-parameter_2009}. We used all noiseless, five-dimensional functions.  Each BBOB ``function'' is a set of functions (commonly referred to as ``instances'') with similar fitness landscapes. In accordance with most comparative studies within the BBOB framework, we study the first five instances. For each of these instances, we run 
every configuration five times. This results in $2,\hspace{-1pt}649,\hspace{-1pt}600$ total runs. The budget for these runs are set at $10^4\cdot D$.

While this gives us data for a total of $4,\hspace{-1pt}608$ different configurations, we ended up only using only one third of the configuration space in our adaptive experiments. This is due to the fact that switching between configurations with a different setting for increasing population ((B)IPOP) would possibly create an information deficit, in that we do not know what population size we should have at the moment of switching. To eliminate this uncertainty, we decided not to consider (B)IPOP in our adaptive experiments, i.e., we dropped two options out of three. This leaves us with $1,\hspace{-1pt}536$ configurations to consider.


\section{Adaptive Configurations}
\label{sec:implementing}

Our main research question in this work is the following. For each of the 24 BBOB functions $f$ we aim to estimate from the detailed runtime data of the 1,\hspace{-1pt}536 static configurations the triples $(C_1,C_2,\tau)$ that exhibit the best performance for optimizing $f$. Each triple is interpreted as follows. The first configuration, $C_1$, is run until hitting target value (``splitpoint'') $\tau$. Immediately after the first iteration in which a solution of fitness $\le \tau$ is sampled, we switch to configuration $C_2$ and run this configuration until the target value $10^{-8}$ is reached or the budget exhausted. 

The long term vision of our research is to extend our study to a landscape-aware configuration. To this end, we will combine our data with explanatory landscape analysis, and to build an automated tool to select CMA-ES configurations \emph{on the fly}. Put differently, we aim at extending the \emph{per instance algorithm configuration} approach analyzed in~\cite{BelkhirDSS17} to the modular CMA-ES and towards a non-static selection. Rather than absolute performance gains, our main interests is therefore in identifying the best performing configurations per each function and each phase of the optimization process.

\subsection{Performance Measures}
\label{sec:performancemeasures}

The hitting time for a certain target is defined as follows:
\begin{definition}[Hitting Time]\label{def:ht}
The \emph{hitting time} of a target $\phi$, a configuration $c$ and a run $i$ on a function $f$ (written as $HT(f,c,\phi,i)$) is defined as the first function evaluation during run $i$ of $c$ in which a function value with a difference of at most $\phi$ to the global optimum is observed. 
\end{definition} 

In this paper, we use two different performance measures. The first is the average hitting time (AHT), which is based on the previously defined hitting time as follows:
\begin{definition}[AHT]
The average hitting time of a configuration $c$ for a target $\phi$ 
on function $f$
is defined as: 
\begin{align*}
AHT(f,c,\phi)=\begin{cases}\infty & \textsc{if } \exists i : HT(f,c,\phi,i) = \infty\\
\frac{\sum_i HT(f,c,\phi,i)}{\sum_i 1} & \textsc{otherwise}
\end{cases}
\end{align*}
\end{definition}
This method places an infinite penalty on non-finished runs. This can be beneficial when dealing with relatively few configurations to avoid selecting those which are unreliable. However, when dealing with more difficult functions and more runs, we want to deal with these non-finished runs differently. This can be done using the expected running time (ERT), which is defined as follows:
\begin{definition}[ERT]
The expected running time (ERT) of a configuration $c$ and target $\phi$ on a function $f$ is calculated as follows: 
$$ERT(f,c,\phi)=\frac{\sum_i (F(f,c,\phi,i) \cdot b + HT(f,c,\phi,i))}{\sum_i (1-F(f,c,\phi,i))},$$
where $b$ is the \emph{budget} of each run (i.e., the maximum number of evaluations) and 
$F(f,c,\phi,i)=1$ if run $i$ on function $f$ failed to reach the target $\phi$, while  $F(f,c,\phi,i)=0$ otherwise.
\end{definition}

Note that if each run of a configuration $c$ on function $f$ found a solution $x$ with $f(x)-f_{opt} \le \phi$, then $ERT(f,c,\phi)=AHT(f,c,\phi)$.

\subsection{Selecting Adaptive Configurations}
\label{sec:method_orig}

We gather the AHTs for targets 
$\Phi = \{ 10^{2-(0.2\cdot i)} \; | \; i \in\{0\dots 50\}\}$.
Based on these AHT values, the adaptive configurations suggested in~\cite{van_rijn_ppns_2018_adpative} are chosen as follows:
\begin{itemize}
\item For each configuration $c$, each of the 24 BBOB functions $f$, and each of the 51 target values $\phi$, we calculate the AHT over all 25 runs (5 runs for each of the first five instances).
\item From this data, we determine the best target value $\phi_{\min}$ for which there exists at least one configuration whose 25 runs all reached this target. 
\item For every target value $\phi\in\Phi$ satisfying 
$\phi>\phi_{\min}$ 
we calculate the best configuration before this target, i.e., we select the configuration $c$ for which $AHT(c,\phi)$ is minimized. 
We denote this configuration $C_1$. We then compute the best configuration $c$ from this target until $\phi_{\min}$, which we denote as $C_2$, i.e., $C_2$ is the configuration for which $AHT(f,c,\phi_{\min})-AHT(f,c,\phi_{\min})$ is minimized. 
In~\cite{van_rijn_ppns_2018_adpative}, the theoretical performance (TH for 'theoretical hitting time') is then calculated as $TH(f,C_1,C_2,\phi)= 
AHT(f,C_1,\phi) - AHT(f,C_2,\phi) +AHT(f,C_2,\phi_{\min})$. 
\item From this data we compute the target value $\tau$ for which the overall performance $TH(f,C_1,C_2,\phi)$ is minimized. This gives us the adaptive configuration $(C_1,C_2,\tau)$. We refer to $\tau$ as the `splitpoint' of the adaptive configuration.
\end{itemize}

\begin{figure}
    \centering
    \includegraphics[scale=0.5]{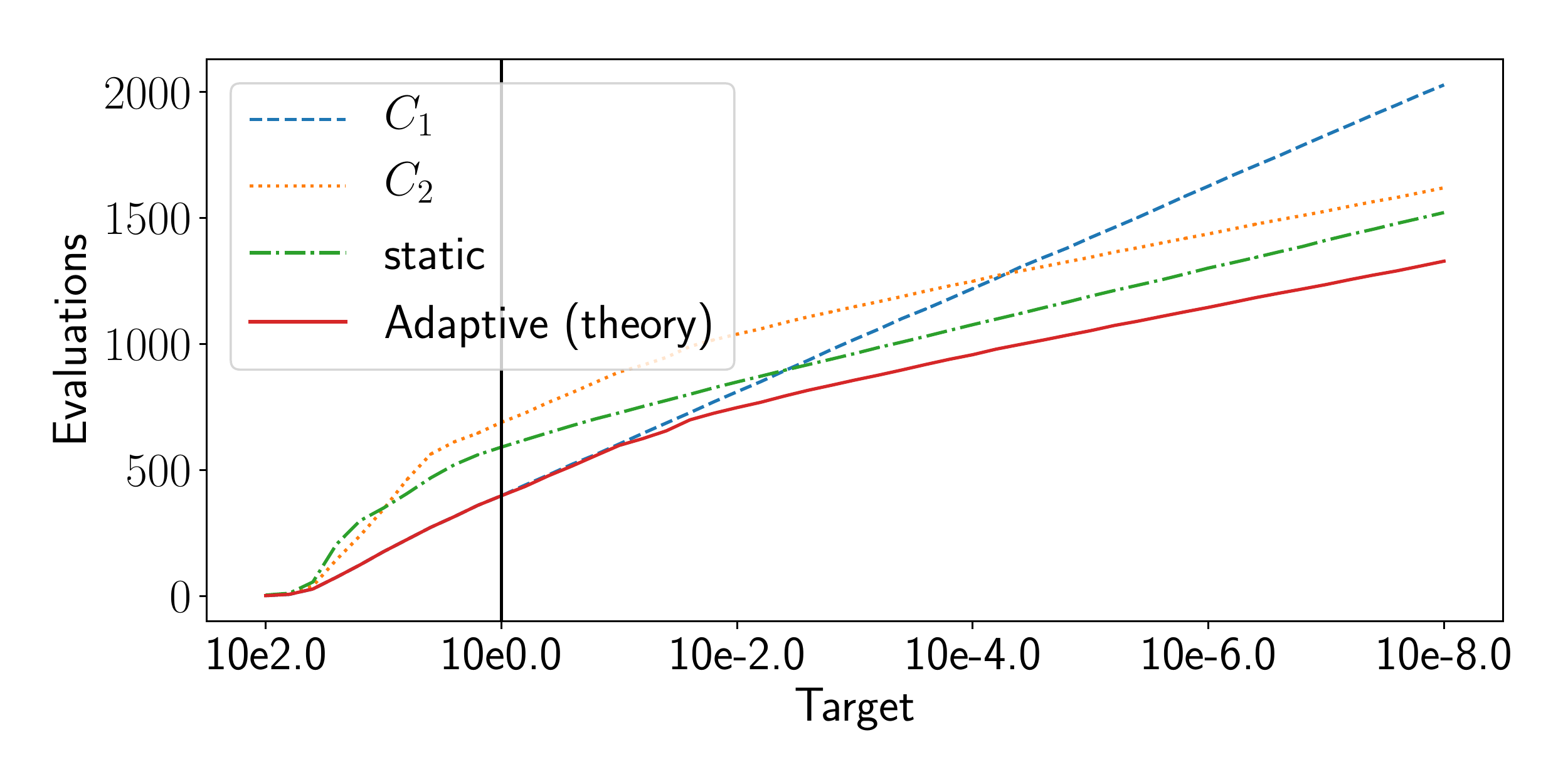}
    \caption{Example of a theoretical adaptive configuration performing better than a static configuration (on F5).
    }
    \label{fig:exampleselection}
\end{figure}

We illustrate this approach in Fig.~\ref{fig:exampleselection}. In this figure we can clearly see that following $C_1$ until the splitpoint and then switching to $C_2$ gives an adaptive configuration which reaches the target with less evaluations than the best static configuration. 

\subsection{Mitigating Uncertainty in the Selection}
\label{sec:alternative}

Initial experiments showed a poor match between the predicted and achieved performance of the adaptive configurations as selected by the previously described procedure. A relatively large variance among the performance profiles of the original runs was identified as cause for this. To select good adaptive configurations despite this variance in the original data, we use the following two techniques:

\textbf{Sliding window}: Instead of considering only the performance at a single target and choosing that as the splitpoint to use, we consider a sliding window around the target, and take the average of the performances at all these targets. 
This should deal with variance by smoothing out the hitting times.
Using this method and the AHT, the value we assign to a splitpoint and configuration pair is defined as follows:

\begin{definition}[Sliding window value]
For a function $f$, adaptive configuration $(C_1,C_2,\phi_i)$ and using $AHT$ as a performance measure, we define the sliding window value $SWV$ as follows:
$$SWV(f,C_1,C_2,\phi_i) = \begin{cases}
\infty & \mkern-74mu \textsc{if } i<w \textsc{ or } (50-i)<w\\
\sum_{j=i-w}^{i+w} HT(f,\phi_j,C_1,C_2) & \mkern18mu \textsc{otherwise}
\end{cases}$$
Here, $w$ determines the radius of the window, in our case we use $w=2$ for a window of size $2w+1=5$. 
\end{definition}

\textbf{Worst Case}: Instead of considering the mean performance, we take a version of the worst-case performance. To still deal with the different instances, we take the worst-case performance for every instance and average these to get a  performance value.
This should reduce the impact of the variance, since even when we assume to get bad runs, this configuration performs the best out of all possible adaptive configurations. This way, the variance might improve the performance of the adaptive configuration.

We run our tests with 3 different settings: sliding window using means, using worstcase, and the original method. 

\subsection{Two-Stage Configuration Selection}
\label{sec:robust}
In addition to these two techniques, we introduce an entirely new procedure to more robustly select which adaptive configurations to run. This is based on the finding that the static configurations are not quite stable enough to use as a baseline. The first step in this process consists of selecting some static configurations for which we should gather more data.
The configurations we will consider are made up of two parts. The first part consists of the 50 best performing static configurations.\footnote{The best static configurations are determined by their $AHT$ at the final reached target. If fewer than 50 configurations reached this target for a function, we extend these configurations by the ones that have the lowest $AHT$ for the previous target. We repeat this process until we have selected 50 configurations.} We then extend this set by looking at the configurations which have been selected to be a part of the 50 theoretically best adaptive configurations. Since this might not be a diverse set of configurations, as one configuration might be chosen as $C_1$ 50 times, we decide to limit the amount of times a certain configuration can be selected as $C_1$ and as $C_2$ to three times each ('limited selection method'). This should give us a more diverse set of configurations which might contribute to good adaptive configurations. We then rerun these configurations using 50 runs on each of the 5 instances, for a total of 250 runs each.

\begin{figure}
    \centering
    \includegraphics[scale=0.5]{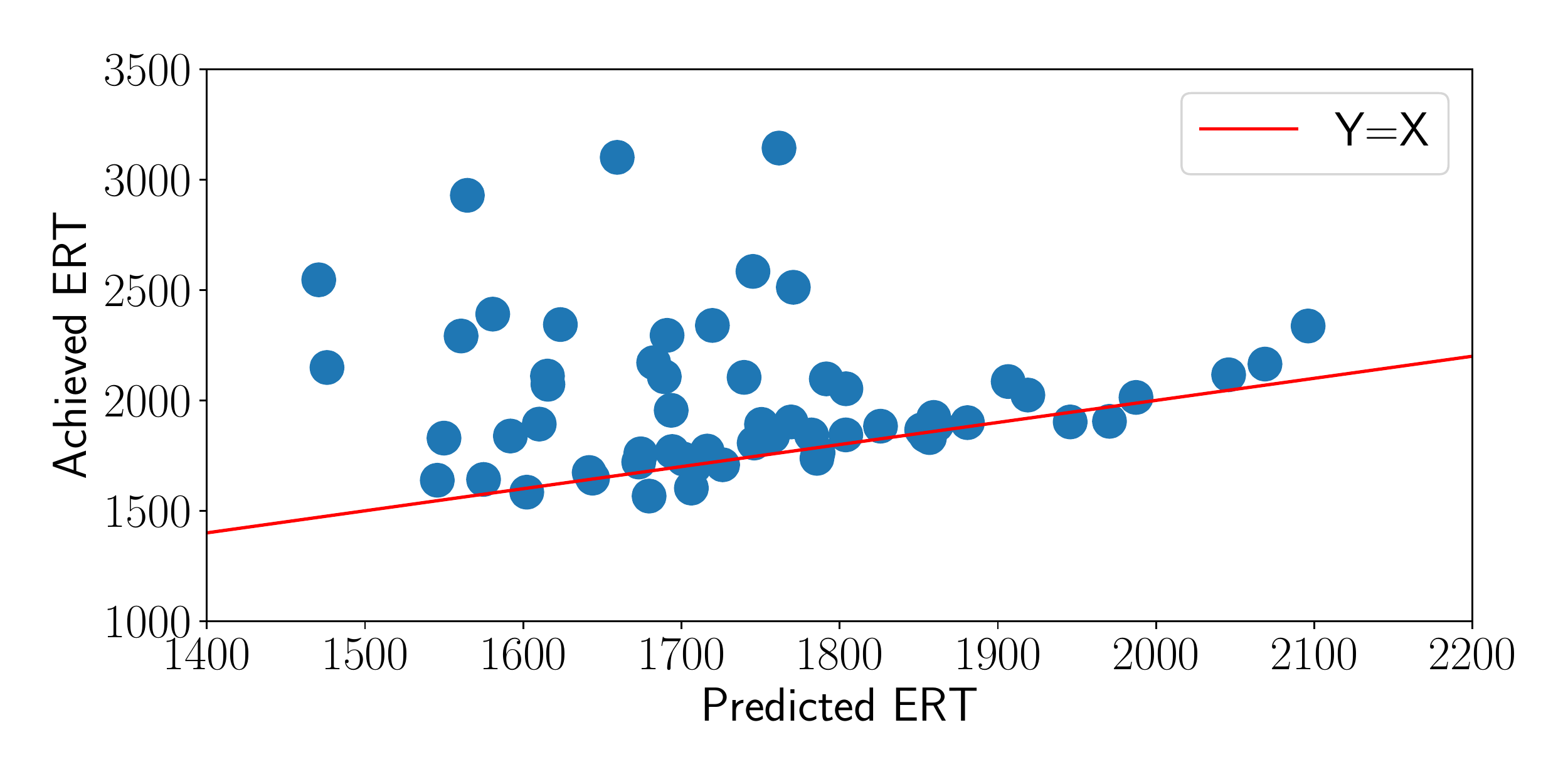}
    \caption{ERT of the selected static configurations for F10. The predicted values are based on the original $5\times 5$ runs, while the achieved ones are on $5\times 50$ runs.}
    \label{fig:expectedachieved}
\end{figure}

In total, this set of configurations will contain anywhere from 50 to 150 configurations to rerun. We can then compare the ERT of the reruns to the original data. This is done for F10 in Fig.~\ref{fig:expectedachieved}, from which we can see that this fit is indeed not very good.

Based on the new, more robust data set, we can now determine the best adaptive configurations to run. This is done using the means and a sliding window method. We still use a sliding window, but instead of taking the worst-case runs for every instance, we revert back to using the AHT. This is done because we have a lot more data available, and the mean should be more stable. We still use the same limited selection method as described previously. 

\section{Initial experiments}
\label{sec:experiments1}
As previously mentioned, our first aim was to extend the work presented in~\cite{van_rijn_ppns_2018_adpative}. We recall that in~\cite{van_rijn_ppns_2018_adpative} only 4 selected functions have been considered. We extend this to all 24 BBOB-functions. We then run the adaptive configurations suggested by this approach, and observe that the fit between the theoretical performance prediction and the actual running times is not very stable.

Using the approach mentioned in the Section~\ref{sec:method_orig}, we can calculate the theoretical improvements when using adaptive configurations. These results are not shown here, but are available in~\cite{data}.
Note that our results differ slightly from those found in the original paper~\cite{van_rijn_ppns_2018_adpative}. 
The reason for this is twofold: First, the configuration space we consider is more restricted since we do not use the (B)IPOP module. Second, we added another restriction on which configurations to use in the adaptive configurations. We require both $C_1$ and $C_{2}$ to reach the target $\phi_{\min}$. This is done to avoid situations where a $C_1$ is selected which very quickly converges to a local optimum and gets stuck there. In practice, switching to $C_{2}$ would then still result in being stuck in this local optimum, which would not be a benefit to finding the global optimum.  

Overall, we found expected improvements to range from $0$ to $15\%$. The $0\%$ improvements are caused by the fact that for some functions, only one or two configurations manage to reach the selected target value. So in these cases, we have $C_1=C_{2}$, leading to the same performance for the adaptive and static configuration. This is present in 4 out of the 24 functions, while all others do have some level of expected improvement.



When running the adaptive configurations,\footnote{Implementation detail: For all our experiments, when switching between different configurations, we reset all static parameters that depend on the configuration. The dynamic parameters such as population, covariance matrix, etc. are kept from the first configurations.} we decided on running them on the same $5$ instances as the original data, but instead of the $5$ runs per instance which we used to gather the data for the static configurations, we run them $50$ times per instance. This should give us more robust results to work with. 

\begin{figure}
    \centering
    \includegraphics[scale=0.5]{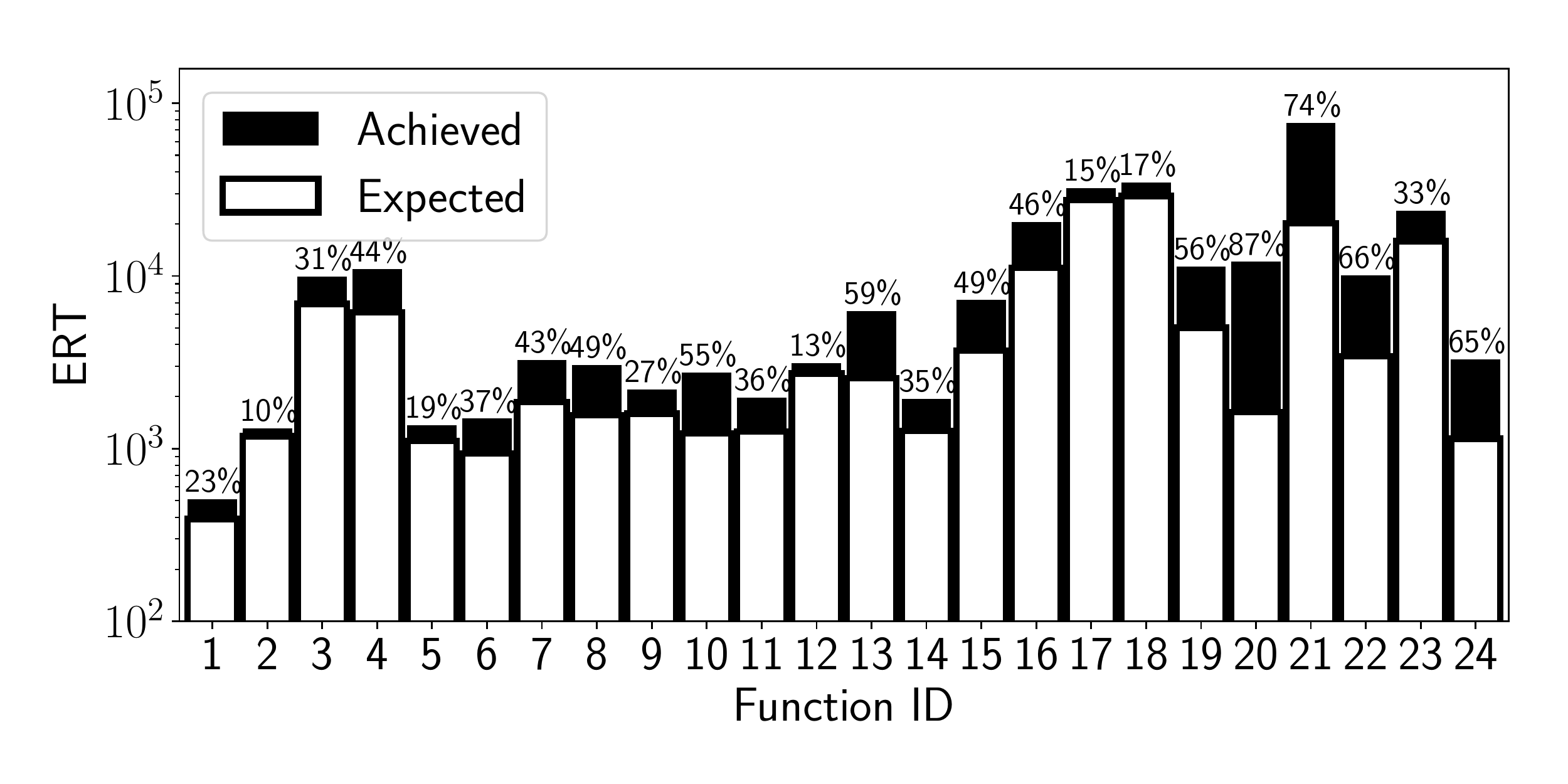}
    \caption{Comparison of the expected ERTs (logscale) for the adaptive configurations against the achieved values. Percentage differences are shown as well.}
    \label{fig:mismatch_opt2}
\end{figure}

The resulting ERTs are compared to the ones we expected in Fig.~\ref{fig:mismatch_opt2}. This comparison shows a clear mismatch between the actually achieved and the expected ERT values. This indicates that the original approach suggested in~\cite{van_rijn_ppns_2018_adpative} and summarized in Section~\ref{sec:method_orig} is not very reliable. 

\section{Results}
\label{sec:results}
We next show how the ideas presented in Sections~\ref{sec:alternative} and~\ref{sec:robust} yield more robust results. 

In a first step, we ran the adaptive configurations determined by the additional methods described in Section~\ref{sec:alternative}. The results from running these adaptive configurations showed that these ideas alone are not sufficient to obtain reliable performance predictions. 
More detailed information about these experiments can be found in~\cite{data}. We summarize the main findings. 

In total, we manage to see improvements (relative to the rerun static configuration) on 10, 8 and 13 functions for the original, sliding window with AHT, and sliding window using worst-case methods, respectively. That is, the sliding window using the  worst-case approach performs slightly better than the other two methods we tested. We expect this to be caused by the fact that this method is less prone to major decreases in performance as a result of variance. 

Even though we do see improvements for several functions, all methods have significant outliers, both in positive and negative sense. This seems to be caused by the fact that the ERT of the static configurations is not robust. As shown in Fig.~\ref{fig:expectedachieved}, the ERT on the $5\times 50$ runs can deviate significantly from the original ERT calculated based on $5\times 5$ runs. And since the static configurations do not have robust ERTs, the adaptive configurations which are based on these values are not necessarily reliable. This lead us to implementing the two-stage approach, as described in Section~\ref{sec:robust}.

\subsection{Comparison of different methods}

\begin{figure*}
    \centering
    \includegraphics[width=\linewidth]{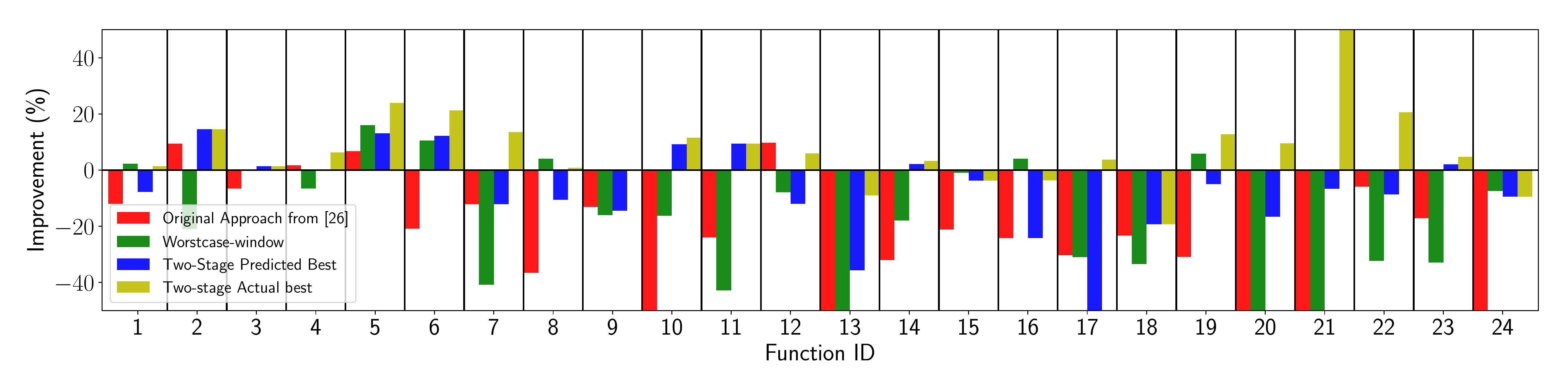}
    \caption{Relative improvement in ERTs of the adaptive configurations over the best static configuration.}
    \label{fig:comp_3methods}
\end{figure*}

We next compare the resulting ERT using three different methods: the original method from Section~\ref{sec:method_orig}, the worst-case method using sliding window from Section~\ref{sec:alternative} and the two-stage method from Section~\ref{sec:robust}. Since the two-stage approach selects and evaluates 50 configurations instead of just one, we show both the ERT of the configuration with the best predicted performance, as well as the one which performed best out of all 50 selected configurations. This is shown in Fig.~\ref{fig:comp_3methods}. From this figure, we can see that for most functions, an improvement relative to the static configuration is achieved for at least some configuration selected in the two-stage method. However, this does not always correspond to the configuration which was predicted to perform best. In the remainder of this paper, we focus on the two-stage approach to see how these performances can be achieved.

\begin{figure}
    \centering
    \includegraphics[scale=0.5]{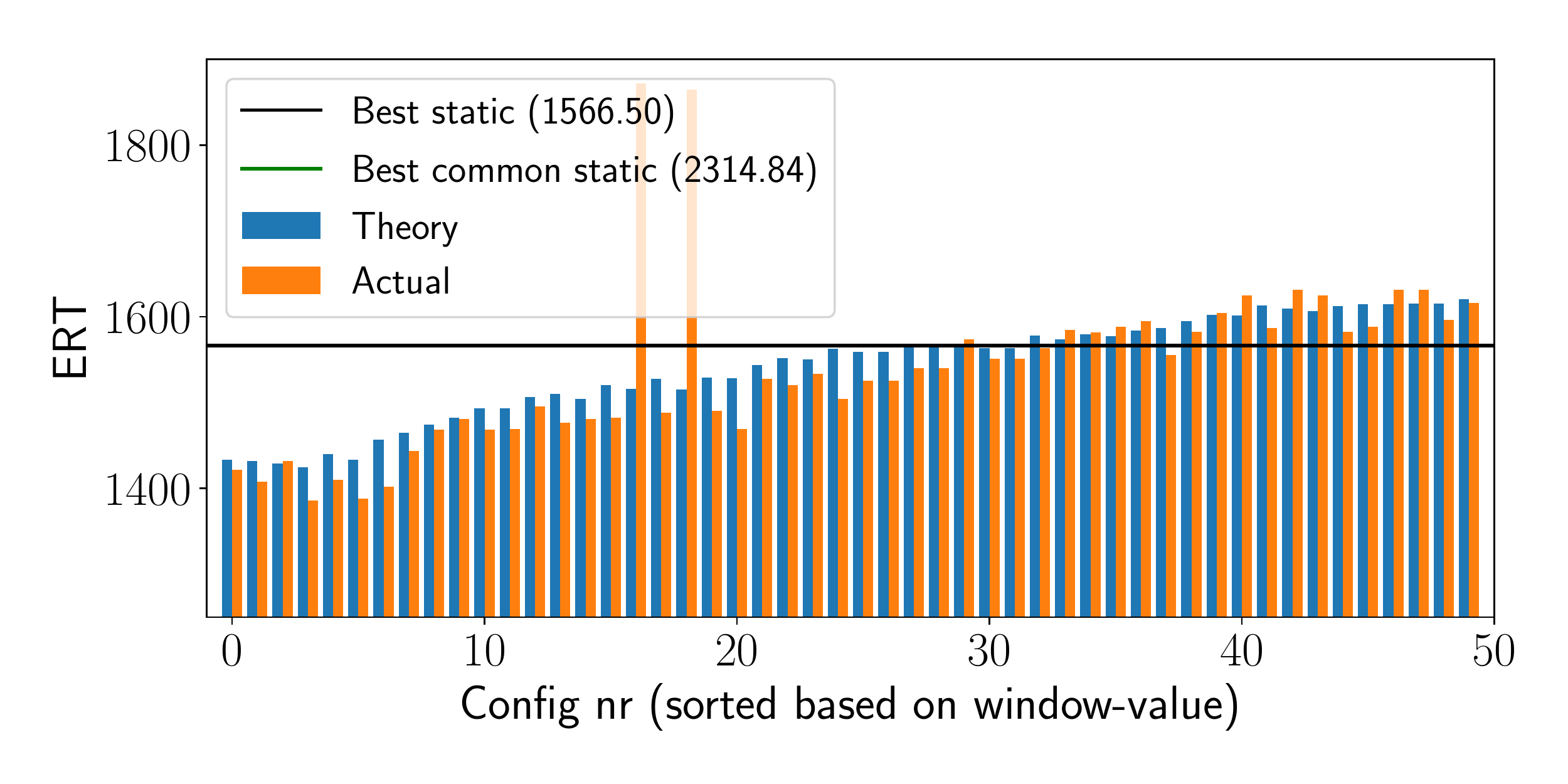}
    \caption{F10: ERT of adaptive configurations compared to the best static and ``common'' static configurations.}
    \label{fig:ert10}
\end{figure}

\begin{table}
    \footnotesize
    \centering
    \begin{tabular}{@{} lll @{}}
        \toprule
        Variant                     & Representation & Short name \\
        \midrule
        CMA-ES                      & 00000000000 & $CM_0$   \\
        Active CMA-ES               & 10000000000 & $CM_1$   \\
        Elitist CMA-ES              & 01000000000 & $CM_2$   \\
        Mirrored-pairwise CMA-ES    & 00100001000 & $CM_3$   \\
        IPOP-CMA-ES                 & 00000000001 & $CM_4$   \\
        Active IPOP-CMA-ES          & 10000000001 & $CM_5$   \\
        Elitist Active IPOP-CMA-ES  & 11000000001 & $CM_6$   \\
        BIPOP-CMA-ES                & 00000000002 & $CM_7$   \\
        Active BIPOP-CMA-ES         & 10000000002 & $CM_8$   \\
        Elitist Active BIPOP-CMA-ES & 11000000002 & $CM_9$   \\
        \bottomrule
    \end{tabular}
    \caption{Common CMA-ES Variants. A selection of ten common CMA-ES variants is listed here, taken from~\cite{van_rijn_evolving_2016}.}
        \label{tab:commons}
\end{table}

\subsection{Performance Comparison}
\label{sec:results-our}
The results of the two-stage method are shown in more detail in Fig.~\ref{fig:ert10} for F10. From this figure, we can see that the fit between theory and practice is quite good, and many of the adaptive configurations manage to outperform the best static configuration by around 10\%. Some outliers are present, but the general trend is positive.
In this figure, we also note the ERT of the best ``common'' CMA-ES variant. 
The set of these configurations is taken from~\cite{van_rijn_evolving_2016} and shown in Table~\ref{tab:commons}. 


Table~\ref{tab:classify} summarizes the results of the two-stage method for all 24 benchmark functions.
The first notable result from this table is the fact that the best ``common'' static configuration can outperform the general best static. This is caused by the fact that these common configurations can have (B)IPOP enabled, which is not the case for the best static. In these cases, we assume that this (B)IPOP module is important to finding the optimum, and an adaptive configuration without this module will not be able to perform very well.

Next, we consider the functions for which the best static ERT is lower than that of the common variants. For these functions, we manage to improve upon this best static configuration when using an adaptive configuration. More specifically, we can see that when the best static configuration from the entire configuration space does not have (B)IPOP enabled, we can reliably achieve an improvement when using adaptive configurations.

We also note that when the best static configuration with (B)IPOP significantly outperforms the best rerun configuration, we do not manage to get the same improvements. If we would consider the best static configurations to include those with (B)IPOP and compare the performance of the adaptive configurations to those, no improvement is made at all. 

In total, we find performance gains on 18 out of 24 functions of the BBOB benchmark, with stable advantages of up to 23\%.

\subsection{Module Activation Plots}
\label{sec:module-activation}

We will now study two functions in more detail. Plots and values for other functions are available in the appendix. 
The functions we will analyze are F10, for which we see a decent improvement for most adaptive configurations, and F24, for which we see very negative results. 

First, we look at which static configurations have been selected, and how they are used within the adaptive configurations. To do this, we introduce what we call \emph{combined module activation plots}. These plots consist of two parts, corresponding to $C_1$ and $C_2$ respectively. In each of these subplots, every line indicates a configuration. The lowest line corresponds with the theoretically best adaptive configuration, increasing from there.

\begin{figure}
    \centering
    \includegraphics[scale=0.5]{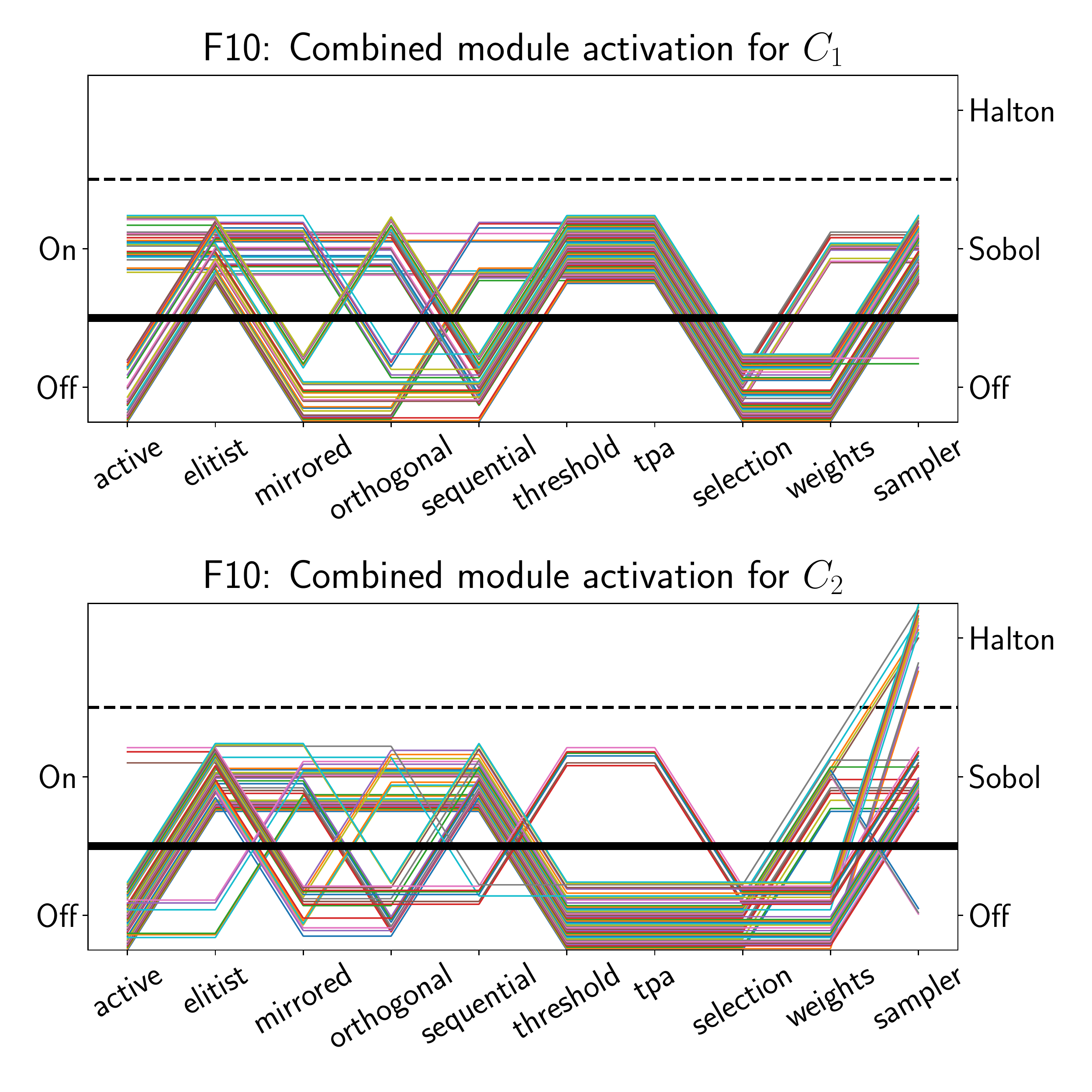}

    \caption{Combined module activation plot for the 50 best adaptive configurations for F10.}
    \label{fig:f10cma}
\end{figure}

\begin{figure}
    \centering
    \includegraphics[scale=0.5]{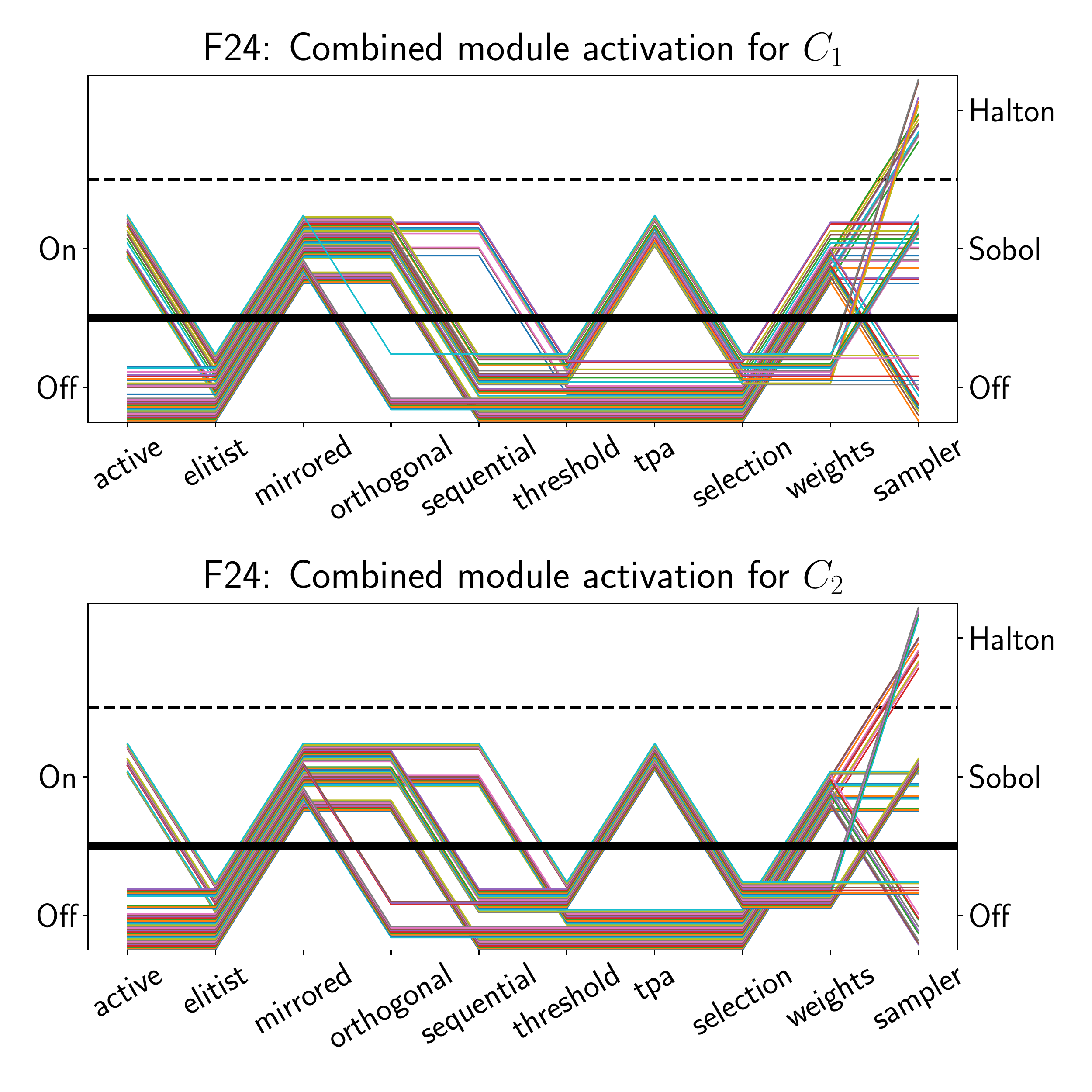}

    \caption{Combined module activation plot for the 50 best adaptive configurations for F24.}
    \label{fig:f24cma}
\end{figure}

In Fig.~\ref{fig:f10cma} and~\ref{fig:f24cma} we see these combined module activation plots for the selected adaptive configurations for F10 and F24 respectively. These figures clearly show that for F10 there is a pattern present among the adaptive configurations: the modules TPA and threshold start activated and in almost all cases get turned off after the splitpoint. Such patterns are not present in the adaptive configurations for F24. This seems to indicate that for F24 the splits are mostly chosen because of small variances between the different configurations, instead of actual inherent properties of the configurations to perform well at certain points of the search. 

\begin{figure*}
    \centering
    \includegraphics[trim={0 0 3cm 3cm},clip,width=\textwidth]{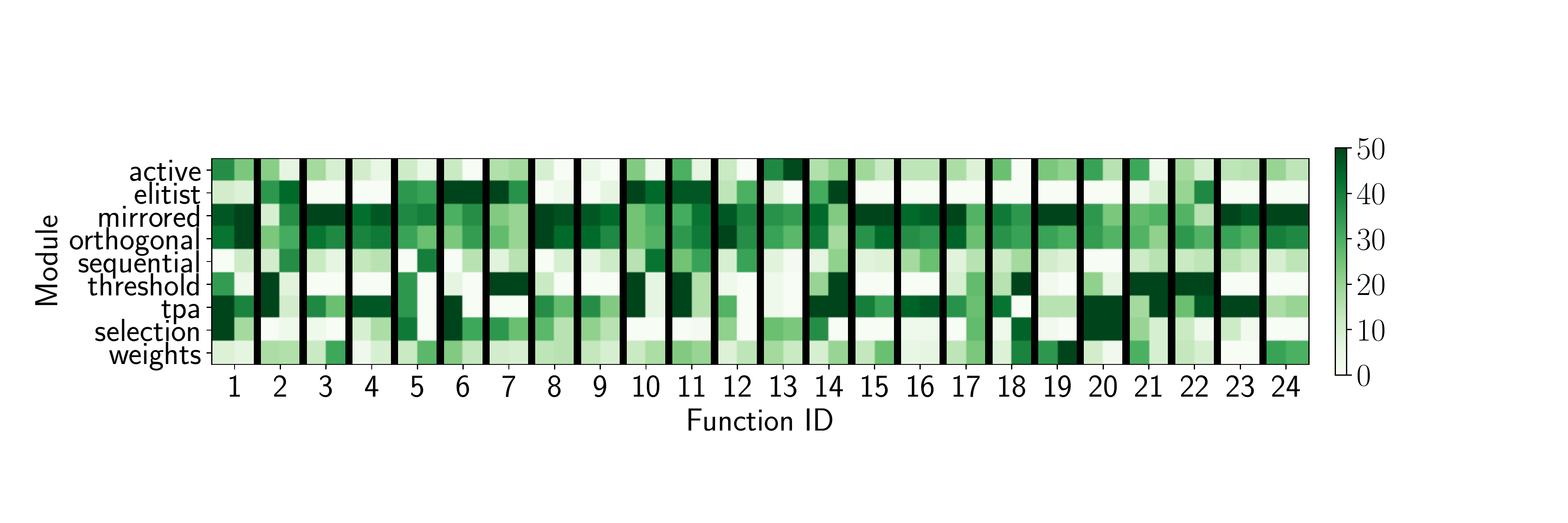}
    \caption{Heatmap of the module activations in the 50 best adaptive configuration for all our functions. For each function, activations for $C_1$ are on the left, while those for $C_{2}$ are on the right. The black lines separate the functions from each other.}
    \label{fig:heatmp_modules}
\end{figure*}

For the other BBOB-functions, we gathered the same data. In Fig.~\ref{fig:heatmp_modules} we plot a heatmap of the module activity in the 50 best adaptive configurations for all functions we considered. For every function and all binary modules, we indicate in how many (out of 50) configurations they are selected, while differentiating between $C_1$ (left) and $C_{2}$ (right). From this figure, we can see that for some functions, there are clear differences between the two parts. These functions correspond to those in which the adaptive configurations improved upon the static ones. For the other functions, the differences between the two parts are not as pronounced. 

\begin{figure}
    \centering
    \includegraphics[scale=0.5]
    {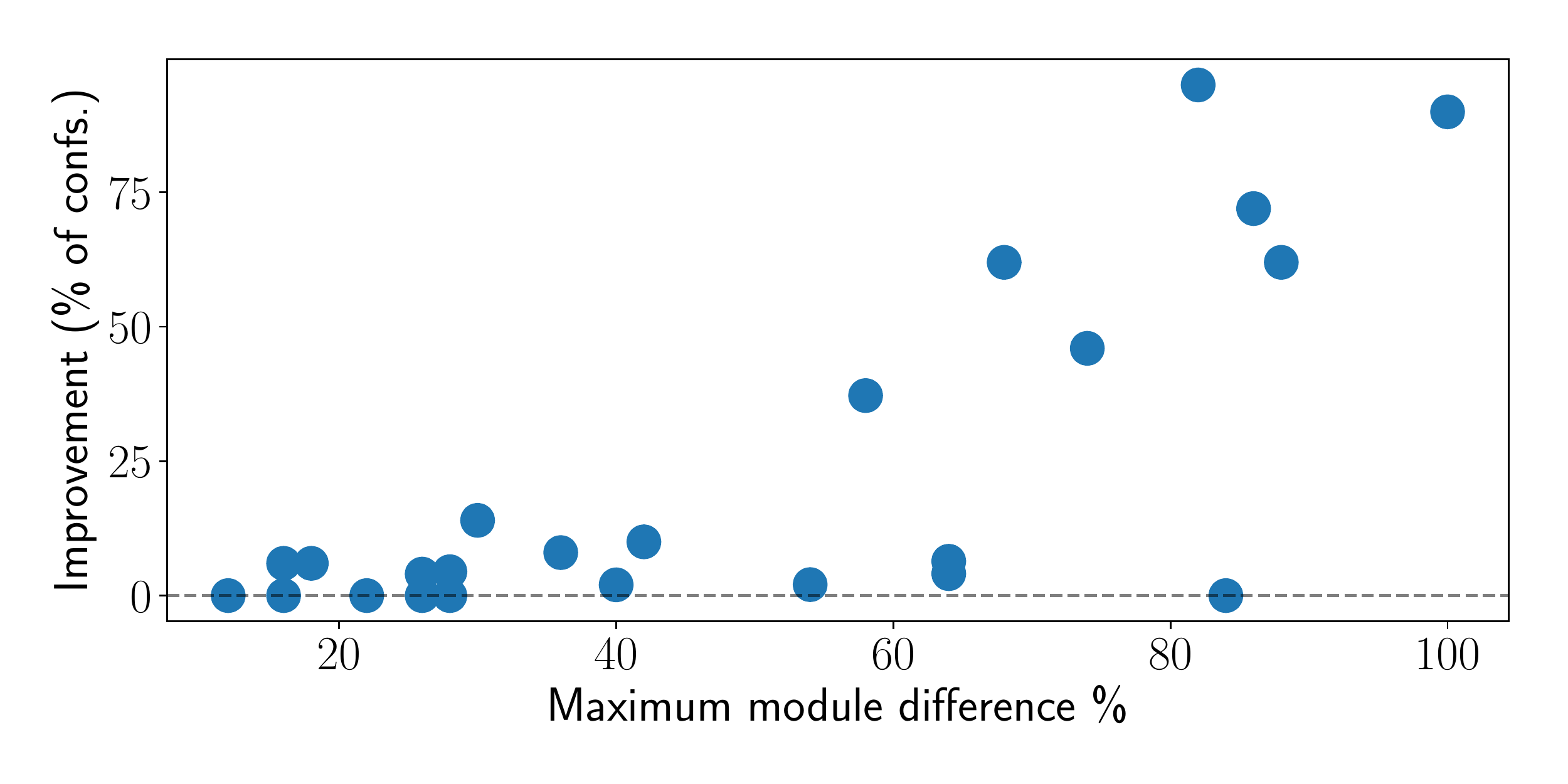}
    \caption{Relation between the maximum module difference \% and the fraction of configurations which manage to improve relative to the best static configuration.}
    \label{fig:module_diff}
\end{figure}

This conclusion is supported by Fig.~\ref{fig:module_diff}, where we show the relation between module difference and the fraction of adaptive configurations which do improve upon the best static configuration. The module difference is calculated as follows: 
$$M = \max_{j=1}^9 \frac{\sum_{i=1}^{50}|C_1(i,j) - C_2(i,j)|}{50} $$
Here, we have $C_1(i,j)$ indicating the value of module $j$ of the configuration chosen as $C_1$ in adaptive configuration number $i$, and the same holds for $C_2(i,j)$. We only use the binary modules in this calculation, so the sampler is ignored.
Based on this figure, we can see that there seems to be a correlation between the module difference percentage and the potential of the adaptive configurations. This supports our claim that the functions where few modules are consistently changed do not exploit any inherent properties of the functions, and instead rely on small differences in search behaviour to predict improvements. While for the functions which do see larger module differences between the first and second part, the predicted improvements are more likely to be based on inherent properties of the function.

\begin{figure*}
    \centering
    \includegraphics
    [width=0.9\linewidth]{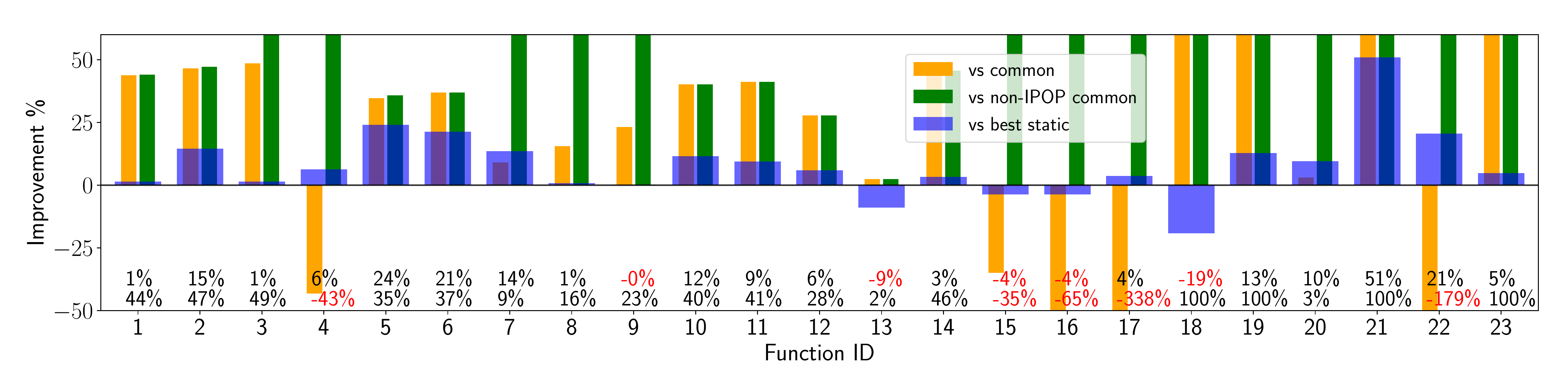}

    \caption{Comparison of the best achieved split ERT relative to the ERTs of the common statics (with and without IPOP; $5\times5$ runs) and the best non-IPOP static on $5\times 50$ runs. Improvements are cut of at 60\% and -50\%, respectively. The precise values of the improvements are shown above the x-axis for the improvement relative to the best static (top) and relative to the best common (bottom) configuration. The precise ERT values can be found in Table~\ref{tab:classify}}
    \label{fig:performance_comparison}
\end{figure*}

\subsection{Summary of Results}
From our experiment, we found large differences  in the potential of our approach between functions. For some functions, such as F10, our approach seems quite stable, resulting in improvements of over $10\%$ for several adaptive configurations, as can be seen in Fig.~\ref{fig:ert10}. However, this is not representative of all functions. We can see from Table~\ref{tab:classify}, in the column ``Amount of impr (\% of configs)'', that there are many functions for which few if any adaptive configurations manage to outperform the static configurations. 

The total improvements gained over the static configurations can be seen in Fig.~\ref{fig:performance_comparison}. In this figure, we can see the relative differences between the best static and best adaptive configurations. We can see that, for the functions where the non-IPOP common static configuration performs similarly to the general best common static ones, the adaptive configuration manages to outperform the static one. This is in line with our previous comments, indicating that the lack of (B)IPOP modules in our configuration space is a large part of the reason why some functions do not see the amount of improvement we would have hoped for.

\begin{figure}
    \centering
    \includegraphics[scale=0.5]{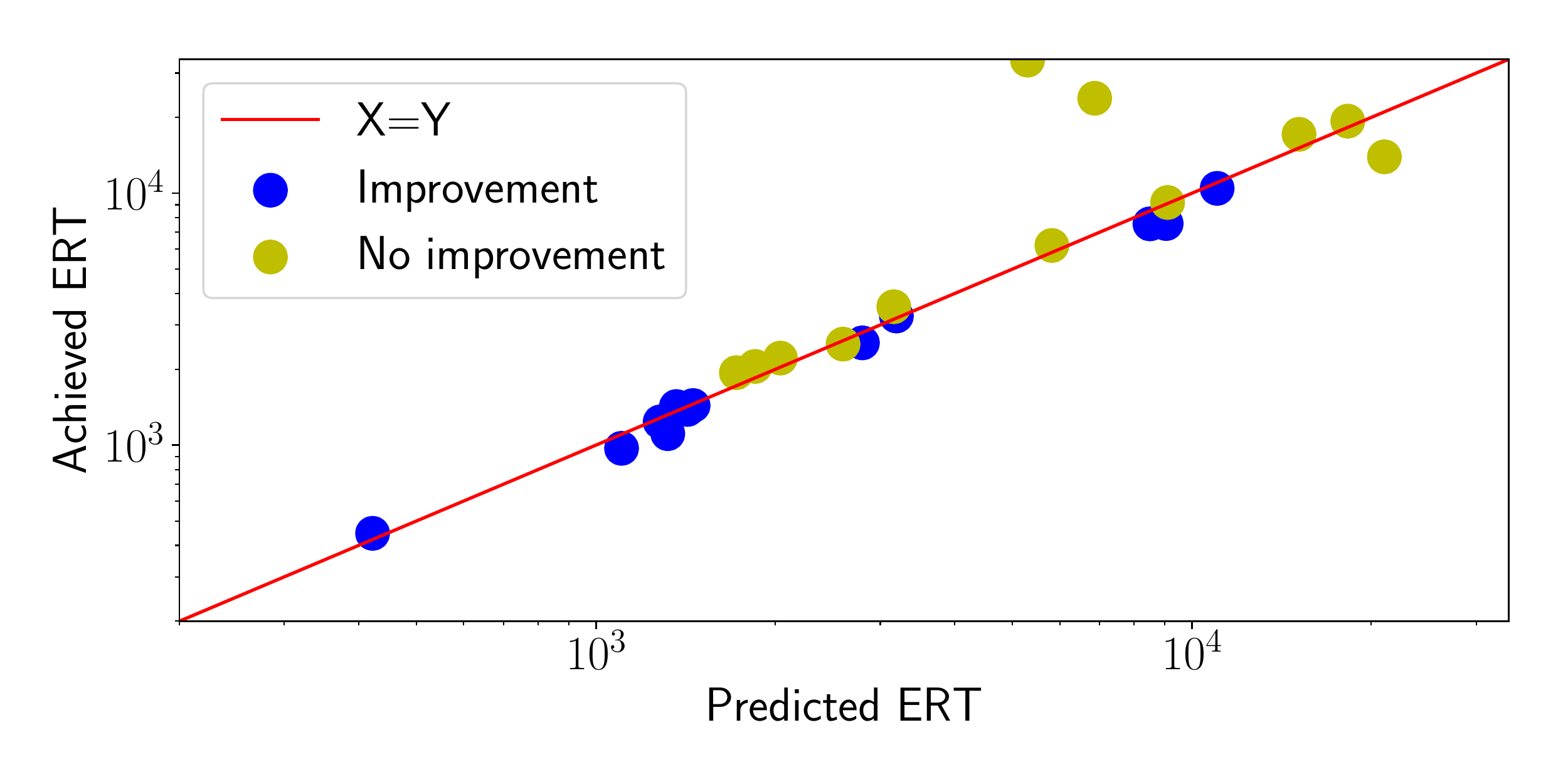}
    \caption{Relation between the best predicted and best achieved ERT for adaptive configurations. Blue dots represent functions for which the average of the 10 best achieved ERT is lower than the best static ERT, while yellow dots indicate the other functions. }
    \label{fig:scatter_imprs}
\end{figure}

To verify that the negative improvements are indeed not caused by our approach itself, we we compare the best expected and achieved ERT of the adaptive configurations for each function. This is shown in Fig.~\ref{fig:scatter_imprs}. We can see that the fit between achieved and predicted ERT is quite good, with some negative outliers for the functions with relatively high predicted ERT. This seems to confirm that our methodology is solid, and that the reason we are not seeing improvement for some functions is most likely caused by the limited configuration space and evaluation budgets.

\section{Conclusion and future work}
\label{sec:conclusion}
In this work, we have shown the potential for configuration adaptation in CMA-ES to outperform static configurations. We have determined that for more difficult functions (B)IPOP might be required to be able to achieve improvements, but for the easier functions this module is not required. 

Since this work breaks the black-box assumption of optimization by including knowledge of the value of the global optimum, which we need to determine when the target $\tau$ is hit to switch configurations, it is not directly applicable to real-world problems.

A next step would be to implement some form of landscape analysis and base the splitpoint and configurations on features of the function. We would expect the improvement on real-world problems to be larger than those we achieved on the BBOB-suite of functions, since we expect that the transition in the local fitness landscape as seen by the algorithms is more drastic than in the ``sterile'' BBOB functions. A cross-validation of our approach on problems found in practice forms  therefore another step that we plan to work on in the near future. 

Another approach would be to extend the framework we used with more modules, and allow the (B)IPOP module to be used in adaptive configurations. This would greatly increase the configuration space, and might necessitate a different approach to determining which configurations to consider.

Finally, we plan on extending our work by tuning the hyper-parameters of the most successful CMA-ES configurations. 

\subsection*{Acknowledgments}
This work was supported by the Paris Ile-de-France Region. 
Our experiments were run on the Leiden DSLAB and DAS-4~\cite{das4_paper} servers.

\appendix
\begin{sidewaystable*}[!thbp]
    \centering
    
\begin{tabular}{
    @{}p{0.03\linewidth}
    p{0.05\linewidth}
    p{0.05\linewidth}
    >{\raggedleft\arraybackslash}p{0.05\linewidth}
    p{0.05\linewidth}
    >{\raggedleft\arraybackslash}p{0.05\linewidth}
    p{0.05\linewidth}
    >{\raggedleft\arraybackslash}p{0.05\linewidth}
    p{0.10\linewidth}
    >{\raggedleft\arraybackslash}p{0.05\linewidth}
    >{\raggedleft\arraybackslash}p{0.05\linewidth}
    >{\raggedleft\arraybackslash}p{0.05\linewidth}
    >{\raggedleft\arraybackslash}p{0.05\linewidth}
    >{\raggedleft\arraybackslash}p{0.05\linewidth}
    >{\raggedleft\arraybackslash}p{0.05\linewidth}@{}
}

\toprule
Fid &   Target & Common config & Common  ERT & Common config  (non IPOP) & non IPOP ERT  &  Best static (B)IPOP  & Best static ERT & Best rerun config & Best rerun ERT & Best adaptive ERT & Relative improvement & Predicted adaptive ERT & Avg 10 splits ERT & Fraction of impr. configs (\%)                          \\
\midrule
1   &  10e-8.0 & $CM_6$ &    795   & $CM_2$ &   799    &  0 &   412 &  00110011010 &    453 &    \textbf{446} &  1.4\% &    \textit{422} &    453 &  6\% \\
2   &  10e-8.0 & $CM_4$ &  2,313   & $CM_2$ & 2,338    &  0 & 1,348 &  10010110010 &  1,448 &  \textbf{1,236} & 14.6\% &          1,282  &  1,283 & 72\% \\
3   &  10e0.4 &  $CM_9$ & 17,871   & /      & $\infty$ &  2 & 2,752 &  00110000110 &  9,316 &  \textbf{9,189} &  1.4\% &          9,104  &  9,771 &  2\% \\
4   &  10e0.8 &  $CM_7$ &  7,301   & /      & $\infty$ &  2 & 2,837 &  00110010010 & 11,162 & \textbf{10,456} &  6.3\% & \textit{11,024} & 11,034 &  6\% \\
5   &  10e-8.0 & $CM_4$ &  1,700   & $CM_3$ & 1,731    &  0 & 1,268 &  00111000010 &  1,461 &  \textbf{1,110} & 24.0\% &  \textit{1,320} &  1,128 & 95\% \\
6   &  10e-8.0 & $CM_2$ &  1,540   & $CM_2$ & 1,540    &  0 & 1,106 &  01110001010 &  1,234 &    \textbf{971} & 21.3\% &  \textit{1,104} &  1,008 & 90\% \\
7   &  10e-8.0 & $CM_5$ &  2,768   & /      & $\infty$ &  1 & 1,471 &  01000101010 & 2,912 & \textbf{2,517} & 13.5\% & \textit{2,598} & 3,002 & 4\% \\
8   &  10e-8.0 & $CM_4$ &  2,623   & /      & $\infty$ &  2 & 1,765 &  00110000010 &  2,233 &  \textbf{2,216} &  0.8\% &  \textit{2,039} &  2,239 &  4\% \\
9   &  10e-8.0 & $CM_4$ &  2,524   & /      & $\infty$ &  1 & 1,726 &  00110000020 &  1,938 &          1,940  & -0.1\% &  \textit{1,720} &  2,095 &  0\% \\
10  &  10e-8.0 & $CM_2$ &  2,315   & $CM_2$ & 2,315    &  2 & 1,437 &  01000110010 &  1,566 &  \textbf{1,386} & 11.5\% &          1,425  &  1,423 & 62\% \\
11  &  10e-8.0 & $CM_0$ &  2,444   & $CM_0$ & 2,444    &  0 & 1,399 &  11110110010 &  1,586 &  \textbf{1,436} &  9.4\% &          1,456  &  1,475 & 62\% \\
12  &  10e-8.0 & $CM_2$ &  4,509   & $CM_2$ & 4,509    &  0 & 3,019 &  00110000010 &  3,463 &  \textbf{3,256} &  6.0\% &  \textit{3,192} &  3,312 & 37\% \\
13  &  10e-8.0 & $CM_1$ &  3,632   & $CM_1$ & 3,632    &  2 & 2,798 &  10110001000 &  3,253 &          3,544  & -8.9\% &  \textit{3,161} &  3,736 &  0\% \\
14  &  10e-8.0 & $CM_0$ &  2,619   & $CM_0$ & 2,619    &  1 & 1,329 &  01001110010 &  1,471 &  \textbf{1,423} &  3.3\% &  \textit{1,364} &  1,434 & 46\% \\
15  &  10e0.4  & $CM_4$ &  4,598   & /      & $\infty$ &  1 & 1,847 &  00110000110 &  5,980 &          6,204  & -3.7\% &          5,822  &  6,883 &  0\% \\
16  &  10e-2.0 & $CM_8$ & 10,379   & /      & $\infty$ &  1 & 3,151 &  00110010010 & 16,534 &         17,140  & -3.7\% & \textit{15,127} & 18,753 &  0\% \\
17  &  10e-4.4 & $CM_7$ &  5,437   & /      & $\infty$ &  2 & 2,209 &  00110010010 & 24,738 & \textbf{23,824} &  3.7\% &  \textit{6,870} & 28,991 &  2\% \\
18  &  10e-4.0 & /      & $\infty$ & /      & $\infty$ &  1 & 6,850 &  00110101110 & 28,312 &         33,754  &-19.2\% &          5,298  & 38,432 &  0\% \\
19  &  10e-0.6 & /      & $\infty$ & /      & $\infty$ &  1 & 3,994 &  00100000120 &  8,666 &  \textbf{7,559} & 12.8\% &  \textit{8,503} &  8,378 & 14\% \\
20  &  10e0.2  & $CM_6$ &  2,628   & /      & $\infty$ &  2 &   769 &  00010011010 &  2,817 &  \textbf{2,547} &  9.6\% &  \textit{2,800} &  2,785 &  8\% \\
21  &  10e-0.6 & /      & $\infty$ & /      & $\infty$ &  2 & 5,889 &  00001110010 & 28,386 & \textbf{13,949} & 50.9\% & \textit{21,040} & 30,338 &  4\% \\
22  &  10e0.0  & $CM_7$ &  2,714   & /      & $\infty$ &  2 &   653 &  00000110010 &  9,546 &  \textbf{7,581} & 20.6\% &  \textit{9,054} &  9,085 & 10\% \\
23  &  10e-0.8 & /      & $\infty$ & /      & $\infty$ &  2 & 5,721 &  00110010010 & 20,296 & \textbf{19,328} &  4.8\% & \textit{18,263} & 21,386 &  6\% \\
24  &  10e1.0  & $CM_8$ &  2,953   & /      & $\infty$ &  2 & 1,187 &  00110000110 &  1,877 &          2,053  & -9.4\% &  \textit{1,850} &  2,358 &  0\% \\
\bottomrule
\end{tabular}

    \caption{Overview of the results from the experiment described in Section~\ref{sec:robust}. The ``Common config'' and ``Common config (non IPOP)'' are selected from the configurations in Table~\ref{tab:commons}.
    The column ``Best static (B)IPOP'' and its ERT (``Best static ERT'') are selected from the full configuration space using the $5\times 5$ run data. The remaining ERTs are from the $5\times 50$ runs, as described in Section~\ref{sec:robust}.
    The columns ``Best adaptive'' and ``Predicted adaptive ERT'' do not necessarily relate to the same configuration. When this is not the case, it is indicated by the italic values in the predicted column. The column ``Relative improvement'' shows the relative improvement in ERT of the best adaptive configuration vs the best rerun static configuration.
    The column ``Avg 10 splits ERT'' is calculated as the average of the 10 best achieved ERTs of the adaptive configurations. The column ``Fraction of impr. configs (\%)'' indicates how many of the adaptive configurations performed better than the ``Best rerun'' configuration.  }
    \label{tab:classify}
\end{sidewaystable*}


\begin{thebibliography}{PMEVBR{\etalchar{+}}15}

\bibitem[ABH11]{auger_mirrored_2011}
Anne Auger, Dimo Brockhoff, and Nikolaus Hansen, \emph{Mirrored {Sampling} in
  {Evolution} {Strategies} with {Weighted} {Recombination}}, Proceedings of the
  13th {Annual} {Conference} on {Genetic} and {Evolutionary} {Computation} (New
  York, NY, USA), {GECCO} '11, ACM, 2011, pp.~861--868.

\bibitem[AH05]{auger_restart_2005}
A.~Auger and N.~Hansen, \emph{A restart {CMA} evolution strategy with
  increasing population size}, 2005 {IEEE} {Congress} on {Evolutionary}
  {Computation}, vol.~2, September 2005, pp.~1769--1776 Vol. 2.

\bibitem[AJT05]{auger2005quasi_random}
Anne Auger, Mohamed Jebalia, and Olivier Teytaud, \emph{Algorithms (x, sigma,
  eta): Quasi-random mutations for evolution strategies}, Artificial Evolution,
  7th International Conference, Evolution Artificielle, {EA} 2005, Lille,
  France, October 26-28, 2005, Revised Selected Papers (El{-}Ghazali Talbi,
  Pierre Liardet, Pierre Collet, Evelyne Lutton, and Marc Schoenauer, eds.),
  Lecture Notes in Computer Science, vol. 3871, Springer, 2005, pp.~296--307.

\bibitem[BAH{\etalchar{+}}10]{brockhoff_mirrored_2010}
Dimo Brockhoff, Anne Auger, Nikolaus Hansen, Dirk~V. Arnold, and Tim Hohm,
  \emph{Mirrored {Sampling} and {Sequential} {Selection} for {Evolution}
  {Strategies}}, Parallel {Problem} {Solving} from {Nature}, {PPSN} {XI},
  Springer, Berlin, Heidelberg, September 2010, pp.~11--21 (en).

\bibitem[BBCPP10]{bartz2010experimental}
Thomas Bartz-Beielstein, Marco Chiarandini, Lu{\'\i}s Paquete, and Mike Preuss,
  \emph{Experimental methods for the analysis of optimization algorithms},
  Springer, 2010.

\bibitem[BBLP10]{bartz2010sequential}
Thomas Bartz-Beielstein, Christian Lasarczyk, and Mike Preuss, \emph{The
  sequential parameter optimization toolbox}, Experimental methods for the
  analysis of optimization algorithms, Springer, 2010, pp.~337--362.

\bibitem[BDSS17]{BelkhirDSS17}
Nacim Belkhir, Johann Dr{\'{e}}o, Pierre Sav{\'{e}}ant, and Marc Schoenauer,
  \emph{Per instance algorithm configuration of {CMA-ES} with limited budget},
  Proc. of Genetic and Evolutionary Computation Conference (GECCO'17), ACM,
  2017, pp.~681--688.

\bibitem[BEdL{\etalchar{+}}16]{das4_paper}
Henri Bal, Dick Epema, Cees de~Laat, Rob van Nieuwpoort, John Romein, Frank
  Seinstra, Cees Snoek, and Harry Wijshoff, \emph{A medium-scale distributed
  system for computer science research: Infrastructure for the long term},
  Computer (2016), no.~5, 54--63.

\bibitem[BFK13]{BFK13}
Thomas B\"ack, Christophe Foussette, and Peter Krause, \emph{Contemporary
  evolution strategies}, Springer, 2013.

\bibitem[BGH{\etalchar{+}}13]{BurkeGHKOOQ13}
Edmund~K. Burke, Michel Gendreau, Matthew~R. Hyde, Graham Kendall, Gabriela
  Ochoa, Ender {\"{O}}zcan, and Rong Qu, \emph{Hyper-heuristics: a survey of
  the state of the art}, {JORS} \textbf{64} (2013), no.~12, 1695--1724.

\bibitem[BMM{\etalchar{+}}07]{burke2007graph}
Edmund~K Burke, Barry McCollum, Amnon Meisels, Sanja Petrovic, and Rong Qu,
  \emph{A graph-based hyper-heuristic for educational timetabling problems},
  European Journal of Operational Research \textbf{176} (2007), no.~1,
  177--192.

\bibitem[HAB{\etalchar{+}}16]{hansen_coco:_2016}
Nikolaus Hansen, Anne Auger, Dimo Brockhoff, Dejan Tušar, and Tea Tušar,
  \emph{{COCO}: {Performance} {Assessment}}, arXiv:1605.03560 [cs] (2016),
  arXiv: 1605.03560.

\bibitem[HAFR09]{hansen_real-parameter_2009}
Nikolaus Hansen, Anne Auger, Steffen Finck, and Raymond Ros,
  \emph{Real-{Parameter} {Black}-{Box} {Optimization} {Benchmarking} 2009:
  {Experimental} {Setup}}, report, INRIA, 2009.

\bibitem[Han08]{hansen_cma-es_2008}
Nikolaus Hansen, \emph{{CMA}-{ES} with {Two}-{Point} {Step}-{Size}
  {Adaptation}}, arXiv:0805.0231 [cs] (2008), arXiv: 0805.0231.

\bibitem[Han09]{hansen_benchmarking_2009}
\bysame, \emph{Benchmarking a {BI}-population {CMA}-{ES} on the {BBOB}-2009
  {Function} {Testbed}}, Proceedings of the 11th {Annual} {Conference}
  {Companion} on {Genetic} and {Evolutionary} {Computation} {Conference}:
  {Late} {Breaking} {Papers} (New York, NY, USA), {GECCO} '09, ACM, 2009,
  pp.~2389--2396.

\bibitem[HAR{\etalchar{+}}10]{hansen2010comparing}
Nikolaus Hansen, Anne Auger, Raymond Ros, Steffen Finck, and Petr
  Po{\v{s}}{\'\i}k, \emph{Comparing results of 31 algorithms from the black-box
  optimization benchmarking bbob-2009}, Proceedings of the 12th annual
  conference companion on Genetic and evolutionary computation, ACM, 2010,
  pp.~1689--1696.

\bibitem[HHLB11]{hutter2011sequential}
Frank Hutter, Holger~H Hoos, and Kevin Leyton-Brown, \emph{Sequential
  model-based optimization for general algorithm configuration}, International
  Conference on Learning and Intelligent Optimization, Springer, 2011,
  pp.~507--523.

\bibitem[HNT16]{hoos2016automated_algorithm_selecion}
Holger~H. Hoos, Frank Neumann, and Heike Trautmann, \emph{Automated algorithm
  selection and configuration (dagstuhl seminar 16412)}, Dagstuhl Reports
  \textbf{6} (2016), no.~10, 33--74.

\bibitem[HO01]{hansen2001self_adaptation_es}
Nikolaus Hansen and Andreas Ostermeier, \emph{Completely derandomized
  self-adaptation in evolution strategies}, Evolutionary Computation \textbf{9}
  (2001), no.~2, 159--195.

\bibitem[JA06]{jastrebski_improving_2006}
G.~A. Jastrebski and D.~V. Arnold, \emph{Improving {Evolution} {Strategies}
  through {Active} {Covariance} {Matrix} {Adaptation}}, 2006 {IEEE}
  {International} {Conference} on {Evolutionary} {Computation}, 2006,
  pp.~2814--2821.

\bibitem[KHNT18]{kerschke2018survey}
Pascal Kerschke, Holger~H. Hoos, Frank Neumann, and Heike Trautmann,
  \emph{Automated algorithm selection: Survey and perspectives}, CoRR
  \textbf{abs/1811.11597} (2018).

\bibitem[KT17]{kerschke2017automated}
Pascal Kerschke and Heike Trautmann, \emph{Automated algorithm selection on
  continuous black-box problems by combining exploratory landscape analysis and
  machine learning}, arXiv preprint arXiv:1711.08921 (2017).

\bibitem[MSKH15]{munoz2015algorithm}
Mario~A Mu{\~n}oz, Yuan Sun, Michael Kirley, and Saman~K Halgamuge,
  \emph{Algorithm selection for black-box continuous optimization problems: a
  survey on methods and challenges}, Information Sciences \textbf{317} (2015),
  224--245.

\bibitem[PMEVBR{\etalchar{+}}15]{piad-morffis_evolution_2015}
A.~Piad-Morffis, S.~Estévez-Velarde, A.~Bolufé-Röhler, J.~Montgomery, and
  S.~Chen, \emph{Evolution strategies with thresheld convergence}, 2015 {IEEE}
  {Congress} on {Evolutionary} {Computation} ({CEC}), May 2015, pp.~2097--2104.

\bibitem[vR18]{modCMA}
Sander van Rijn, \emph{Modular cma-es framework
  from~\cite{van_rijn_evolving_2016}, v0.3.0},
  \url{https://github.com/sjvrijn/ModEA}. Available also as pypi package at
  \url{https://pypi.org/project/ModEA/0.3.0/}, 2018.

\bibitem[VRDB18]{van_rijn_ppns_2018_adpative}
Sander Van~Rijn, Carola Doerr, and Thomas B{\"a}ck, \emph{Towards an adaptive
  cma-es configurator}, International Conference on Parallel Problem Solving
  from Nature, Springer, 2018, pp.~54--65.

\bibitem[vRWvLB16]{van_rijn_evolving_2016}
Sander van Rijn, Hao Wang, Matthijs van Leeuwen, and Thomas B\"ack,
  \emph{Evolving the structure of {Evolution} {Strategies}}, 2016 {IEEE}
  {Symposium} {Series} on {Computational} {Intelligence} ({SSCI}), December
  2016, pp.~1--8.

\bibitem[VvRDB19]{data}
Diederick Vermetten, Sander van Rijn, Carola Doerr, and Thomas B\"ack, 2019,
  GitHub repository containing more data and experiments from this project.

\bibitem[WEB14]{wang_mirrored_2014}
Hao Wang, Michael Emmerich, and Thomas B\"ack, \emph{Mirrored {Orthogonal}
  {Sampling} with {Pairwise} {Selection} in {Evolution} {Strategies}},
  Proceedings of the 29th {Annual} {ACM} {Symposium} on {Applied} {Computing}
  (New York, NY, USA), {SAC} '14, ACM, 2014, pp.~154--156.

\end{thebibliography}

\newcommand{\etalchar}[1]{$^{#1}$}
\providecommand{\bysame}{\leavevmode\hbox to3em{\hrulefill}\thinspace}
\providecommand{\MR}{\relax\ifhmode\unskip\space\fi MR }
\providecommand{\MRhref}[2]{%
  \href{http://www.ams.org/mathscinet-getitem?mr=#1}{#2}
}
\providecommand{\href}[2]{#2}

\end{document}